\documentclass[conference]{IEEEtran}
\IEEEoverridecommandlockouts
\usepackage[utf8]{inputenc}
\usepackage{newunicodechar}
\usepackage[english]{babel}
\usepackage{amsmath,amssymb,amsfonts}
\usepackage{algorithmic}
\usepackage{graphicx}
\usepackage{textcomp}
\usepackage{xcolor}
\usepackage{tikz}
\usepackage{pgfplots}
\usepackage{diagbox}
\usepackage{colortbl}
\usepackage{array}
\usepackage{cite}
\usepackage{hyperref}
\def\BibTeX{{\rm B\kern-.05em{\sc i\kern-.025em b}\kern-.08em
    T\kern-.1667em\lower.7ex\hbox{E}\kern-.125emX}}
\graphicspath{ {./images/} }

\IEEEoverridecommandlockouts
\IEEEpubid{\makebox[\columnwidth]{978-0-7381-4394-1/21/\$31.00~\copyright2021 IEEE\hfill} \hspace{\columnsep}\makebox[\columnwidth]{ }}

\newcommand\copyrighttext{%
  \footnotesize \textcopyright 2021 IEEE. Personal use of this material is permitted. Permission from IEEE must be obtained for all other uses, in any current or future media, including reprinting/republishing this material for advertising or promotional purposes, creating new collective works, for resale or redistribution to servers or lists, or reuse of any copyrighted component of this work in other works.
  DOI: \href{https://doi.org/10.1109/CCWC51732.2021.9375950}{10.1109/CCWC51732.2021.9375950}}
\newcommand\copyrightnotice{%
\begin{tikzpicture}[remember picture,overlay]
\node[anchor=south,yshift=10pt] at (current page.south) {\fbox{\parbox{\dimexpr\textwidth-\fboxsep-\fboxrule\relax}{\copyrighttext}}};
\end{tikzpicture}%
}

\title{CondenseNeXt: An Ultra-Efficient Deep Neural Network for Embedded Systems\\}

\begin{document}
\author{\IEEEauthorblockN{Priyank Kalgaonkar}
\IEEEauthorblockA{\textit{Department of Electrical and Computer Engineering} \\
\textit{Purdue School of Engineering and Technology}\\
Indianapolis, Indiana 46202, USA. \\
pkalgaon@purdue.edu}
\and
\IEEEauthorblockN{Mohamed El-Sharkawy}
\IEEEauthorblockA{\textit{Department of Electrical and Computer Engineering} \\
\textit{Purdue School of Engineering and Technology}\\
Indianapolis, Indiana 46202, USA. \\
melshark@purdue.edu}
}

\maketitle
\copyrightnotice
\IEEEpubidadjcol

\begin{abstract}
Due to the advent of modern embedded systems and mobile devices with constrained resources, there is a great demand for incredibly efficient deep neural networks for machine learning purposes. There is also a growing concern of privacy and confidentiality of user data within the general public when their data is processed and stored in an external server which has further fueled the need for developing such efficient neural networks for real-time inference on local embedded systems. The scope of our work presented in this paper is limited to image classification using a convolutional neural network. A Convolutional Neural Network (CNN) is a class of Deep Neural Network (DNN) widely used in the analysis of visual images captured by an image sensor, designed to extract information and convert it into meaningful representations for real-time inference of the input data. In this paper, we propose a neoteric variant of deep convolutional neural network architecture to ameliorate the performance of existing CNN architectures for real-time inference on embedded systems. We show that this architecture, dubbed CondenseNeXt, is remarkably efficient in comparison to the baseline neural network architecture, CondenseNet, by reducing trainable parameters and FLOPs required to train the network whilst maintaining a balance between the trained model size of less than 3.0 MB and accuracy trade-off resulting in an unprecedented computational efficiency.
\end{abstract}
\vskip 0.06in
\begin{IEEEkeywords}
Deep Neural Network, Convolutional Neural Network, CondenseNeXt, CondenseNet, Depthwise Separable Convolutions, Group-wise Pruning, PyTorch, CIFAR-10.
\end{IEEEkeywords}


\section{Introduction}
Convolutional Neural Networks, a class of Deep Neural Networks, have been gaining popularity in the recent years and developing more efficient state-of-the-art deep neural network architectures has been a subject of significant attention. The first working DNN architecture was published by Alexey G. Ivakhnenko and V. G. Lapa in 1967 for supervised, deep, feed-forward, multi-layer perceptions \cite{b12}. In 1971, Alexey G. Ivakhnenko introduced another DNN architecture with eight layers trained by multi-parametric datasets using the Group Method of Data Handling (GMDH) algorithm \cite{b29}.

Fast forwarding to the 21\textsuperscript{st} century, DNN has become an important part of Computer Vision which is a multidisciplinary scientific field of study that seeks to develop novel techniques to enable computers to perform complex tasks \cite{b14} such as  image classification and object detection, analogous to that of human visual system. Due to the digitization of our society, inexpensive cameras and Internet of Things (IoT), we now have access to a fairly large amount of data for multinomial classification.


\section{Background}
G. Huang {\textit{et al.}} introduced ResNet in 2016 to facilitate the training of deep neural networks with copious amounts of data. This model introduces \textit{identity shortcut connections} that skips one or more layers, a fundamental ground-breaking idea proposed by ResNet  \cite{b14}. Although it is relatively an old DNN architecture, it is still used as a baseline due to it's inventive impact.

In 2017, G. Huang {\textit{et al.}} took the idea of residual connections even further and introduced a new DNN architecture called DenseNet in which each layer is connected to every other layer in a feed-forward fashion. For each layer, the feature-maps of all preceding layers are used as inputs, and its own feature-maps are used as inputs into all subsequent layers \cite{b15}. This  is an ingenious  technique  since  it  facilitates  the  reuse  of channels and neurons (tensors) that represent information at different level of coarseness. 


 In 2018, G. Huang {\textit{et al.}} introduced CondenseNet \cite{b10}, an improvement over DenseNet. CondenseNet replaces the classical convolutions with learned grouped convolutions which is designed to allow the network to learn mapping between channels and groups instead of separating channels on the basis of their offsets which results in different groups of channels being trained separately. With recent developments in the area of embedded systems with constrained computational resources for autonomous vehicles \cite{b81}, robotics \cite{b82} and unmanned ariel vehicles such as drones \cite{b9}, a desire for more efficient yet accurate inferring DNN models has burgeoned in recent years \cite{b4, b28}. In this paper, we propose a deep neural network architecture with a trained model size of less than 3.0 MB to address this demand.


\section{Prior Work}
Prior work in the following areas have contributed to the research work presented within this paper:
\begin{itemize}
\setlength\itemsep{0.35em}
  \item Convolutional neural networks \cite{b23,b4,b33}, specifically CondenseNet \cite{b10}, form the baseline and are similar to our proposed architecture in few aspects which will be discussed in the following sections.
  \item The DenseNet convolutional neural network architecture which first manifested the advantages of dense connectivity in which each layer is connected to every other layer in a feed-forward fashion \cite{b15}. This results in the reuse of channels and neurons (tensors) that represent information at different level of coarseness.
  \item Depthwise separable convolution, which was first seen in \cite{b185}, is one of the main fundamentals of our proposed architecture. Although the idea of utilizing spatially separable convolutions in deep neural networks can be traced back to 2012 \cite{b285}, the depthwise interpretation of spatially separable convolutions is comparatively new and has been proven to be widely successful in its implementation in MobileNet \cite{b27} and Xception \cite{b34} architectures.
  \item Model compression techniques such as \cite{b2,b19,b31} are designed to improve the efficiency of a neural network architecture. We choose to incorporate group-wise filter pruning \cite{b45} process into the training stage which allows the network to train from scratch with an unprecedented computational efficiency.
\end{itemize}
\vskip 0.1in
\section{CondenseNeXt Architecture}
\smallskip
In this paper, we propose a novel CNN architecture inspired by the aforementioned prior works. We have named our neural network architecture \textit{CondenseNeXt} referring to the \textit{next} dimension of cardinality. In this section, we describe the architecture of our proposed neural network in detail.

\subsection{Convolution Layers}

CondenseNet utilizes learned group convolutions technique where inconsequential filters with low magnitude weights are pruned first and then the filters are learned after the groupings are finalized. Grouped convolutions form an underlying fundamental of CondenseNet which is a peculiar case of sparsely connected convolutional layers. Here, the input and output neurons are divided into independent groups of $g$. This enables the network to perform regular spatial convolution over each independent group and concatenate the resulting feature maps. 

The goal of CondenseNeXt is to reduce the trained model size to less than 3.0 MB and also the amount of computational resources required to train the network from scratch. In order to achieve this, we first replace the grouped convolution with depthwise separable convolution in learned group convolution. Depthwise separable convolution is comprised of:
\begin{itemize}
\setlength\itemsep{0.35em}
    \item Depthwise convolution (filtering) layer: Convolution to a single input channel is applied independently whereas in standard convolution, convolution is applied to all input channels. 
    \item Pointwise convolution (combining) layer: A linear combination is carried out for each of these layers.
\end{itemize}

The depthwise separable convolution splits a kernel into two discrete filters for filtering and combining stages respectively as shown in Figure 1 above, which reduces the computation required for training as well as size of the trained model.

Consider a standard convolutional filter $K$ of size $H \times H \times I \times O$ where $I$ is the number of input channels and $O$ is the number of output channels with an input feature map \textbf{$X$} of size $D_{x} \times D_{x} \times I$ that produces an output feature map \textbf{$Y$} of size $D_{x} \times D_{x}\times O$ can be computed as follows:

\begin{equation}
\label{eqn_1}
Y_{k,l,n} = \sum_{i,j,m} k_{i,j,m,n} \cdot X_{k+i-1,l+j-1,m}
\end{equation}

For depthwise separable convolution, (\ref{eqn_1}) can be factorized into two parts: the first part applies a $3 \times 3$ depthwise convolution $\hat{K}$ with one filter for each input channel as follows:

\begin{equation}
\label{eqn_2}
\hat{Y}_{k,l,m} = \sum_{i,j} \hat{K}_{i,j,m} \cdot X_{k+i-1,l+j-1,m}
\end{equation}

And then the second part applies a $1 \times 1$ pointwise convolution $\tilde{K}$ in order to perform linear combination and combine the outputs of depthwise convolution as follows:

\begin{equation}
\label{eqn_3}
Y_{k,l,n} = \sum_{m} \tilde{K}_{m,n} \cdot \hat{Y}_{k-1,l-1,m}
\end{equation}

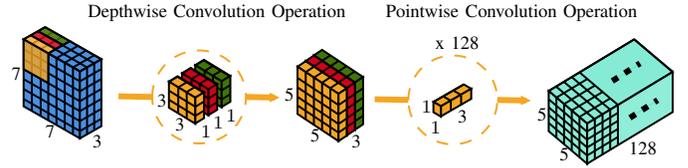
\begin{figure}[t]
\tikzset{every picture/.style={line width=0.75pt}} 

\begin{tikzpicture}[scale=0.51,x=0.75pt,y=0.75pt,yscale=-1,xscale=1]

\draw  [fill={rgb, 255:red, 65; green, 117; blue, 5 }  ,fill opacity=1 ] (202.96,182.02) -- (229.1,197.97) -- (220.49,203.32) -- (194.34,187.37) -- (194.34,187.37) -- cycle ;
\draw  [fill={rgb, 255:red, 65; green, 117; blue, 5 }  ,fill opacity=1 ] (194.34,213.14) -- (194.34,187.38) -- (220.49,203.32) -- (220.49,229.08) -- cycle ;
\draw [fill={rgb, 255:red, 65; green, 117; blue, 5 }  ,fill opacity=1 ]   (211.9,197.68) -- (211.98,223.88) ;
\draw [fill={rgb, 255:red, 65; green, 117; blue, 5 }  ,fill opacity=1 ]   (203.65,192.93) -- (203.47,218.49) ;
\draw  [fill={rgb, 255:red, 65; green, 117; blue, 5 }  ,fill opacity=1 ] (229.1,223.55) -- (220.49,228.73) -- (220.49,203.15) -- (229.1,197.97) -- cycle ;
\draw [fill={rgb, 255:red, 65; green, 117; blue, 5 }  ,fill opacity=1 ]   (220.49,212.28) -- (229.1,207.1) ;
\draw [fill={rgb, 255:red, 65; green, 117; blue, 5 }  ,fill opacity=1 ]   (220.49,221.27) -- (229.1,216.08) ;
\draw [fill={rgb, 255:red, 65; green, 117; blue, 5 }  ,fill opacity=1 ]   (211.9,197.68) -- (220.52,192.49) ;
\draw [fill={rgb, 255:red, 65; green, 117; blue, 5 }  ,fill opacity=1 ]   (203,193.44) -- (211.62,188.25) ;
\draw [fill={rgb, 255:red, 65; green, 117; blue, 5 }  ,fill opacity=1 ]   (194.09,187.87) -- (202.71,182.52) ;
\draw [fill={rgb, 255:red, 65; green, 117; blue, 5 }  ,fill opacity=1 ]   (194.34,205.32) -- (220.49,221.27) ;
\draw [fill={rgb, 255:red, 65; green, 117; blue, 5 }  ,fill opacity=1 ]   (194.34,196.34) -- (220.49,212.28) ;
\draw [fill={rgb, 255:red, 65; green, 117; blue, 5 }  ,fill opacity=1 ]   (202.71,182.52) -- (228.85,198.46) ;

\draw  [fill={rgb, 255:red, 208; green, 2; blue, 27 }  ,fill opacity=1 ] (189.96,190.02) -- (216.1,205.97) -- (207.49,211.32) -- (181.34,195.37) -- (181.34,195.37) -- cycle ;
\draw  [fill={rgb, 255:red, 208; green, 2; blue, 27 }  ,fill opacity=1 ] (181.34,221.14) -- (181.34,195.38) -- (207.49,211.32) -- (207.49,237.08) -- cycle ;
\draw [fill={rgb, 255:red, 208; green, 2; blue, 27 }  ,fill opacity=1 ]   (198.9,205.68) -- (198.98,231.88) ;
\draw [fill={rgb, 255:red, 208; green, 2; blue, 27 }  ,fill opacity=1 ]   (190.65,200.93) -- (190.47,226.49) ;
\draw  [fill={rgb, 255:red, 208; green, 2; blue, 27 }  ,fill opacity=1 ] (216.1,231.55) -- (207.49,236.73) -- (207.49,211.15) -- (216.1,205.97) -- cycle ;
\draw [fill={rgb, 255:red, 208; green, 2; blue, 27 }  ,fill opacity=1 ]   (207.49,220.28) -- (216.1,215.1) ;
\draw [fill={rgb, 255:red, 208; green, 2; blue, 27 }  ,fill opacity=1 ]   (207.49,229.27) -- (216.1,224.08) ;
\draw [fill={rgb, 255:red, 208; green, 2; blue, 27 }  ,fill opacity=1 ]   (198.9,205.68) -- (207.52,200.49) ;
\draw [fill={rgb, 255:red, 208; green, 2; blue, 27 }  ,fill opacity=1 ]   (190,201.44) -- (198.62,196.25) ;
\draw [fill={rgb, 255:red, 208; green, 2; blue, 27 }  ,fill opacity=1 ]   (181.09,195.87) -- (189.71,190.52) ;
\draw [fill={rgb, 255:red, 208; green, 2; blue, 27 }  ,fill opacity=1 ]   (181.34,213.32) -- (207.49,229.27) ;
\draw [fill={rgb, 255:red, 208; green, 2; blue, 27 }  ,fill opacity=1 ]   (181.34,204.34) -- (207.49,220.28) ;
\draw [fill={rgb, 255:red, 208; green, 2; blue, 27 }  ,fill opacity=1 ]   (189.71,190.52) -- (215.85,206.46) ;

\draw  [fill={rgb, 255:red, 80; green, 227; blue, 194 }  ,fill opacity=0.69 ] (628.51,161.8) -- (666.53,194.02) -- (581.93,245.12) -- (543.9,212.91) -- (628.51,161.8) -- cycle ;
\draw  [fill={rgb, 255:red, 80; green, 227; blue, 194 }  ,fill opacity=0.64 ] (581.93,245.12) -- (581.93,284.92) -- (543.9,252.69) -- (543.9,212.9) -- cycle ;
\draw  [fill={rgb, 255:red, 80; green, 227; blue, 194 }  ,fill opacity=0.64 ] (666.53,194.02) -- (666.53,233.79) -- (581.93,284.91) -- (581.93,245.14) -- cycle ;
\draw [fill={rgb, 255:red, 80; green, 227; blue, 194 }  ,fill opacity=0.64 ]   (543.9,221.06) -- (581.93,253.28) ;
\draw [fill={rgb, 255:red, 80; green, 227; blue, 194 }  ,fill opacity=0.64 ]   (543.9,229.02) -- (581.93,261.23) ;
\draw [fill={rgb, 255:red, 80; green, 227; blue, 194 }  ,fill opacity=0.64 ]   (543.9,236.58) -- (581.93,268.8) ;
\draw [fill={rgb, 255:red, 80; green, 227; blue, 194 }  ,fill opacity=0.64 ]   (543.9,244.54) -- (581.93,276.76) ;
\draw [fill={rgb, 255:red, 80; green, 227; blue, 194 }  ,fill opacity=0.64 ]   (559.11,225.83) -- (559.11,265.62) ;
\draw [fill={rgb, 255:red, 80; green, 227; blue, 194 }  ,fill opacity=0.64 ]   (551.51,219.47) -- (551.51,259.25) ;
\draw [fill={rgb, 255:red, 80; green, 227; blue, 194 }  ,fill opacity=0.64 ]   (566.72,232.2) -- (566.72,271.99) ;
\draw [fill={rgb, 255:red, 80; green, 227; blue, 194 }  ,fill opacity=0.64 ]   (574.32,238.56) -- (574.32,278.35) ;
\draw [fill={rgb, 255:red, 80; green, 227; blue, 194 }  ,fill opacity=0.64 ]   (597.14,236.18) -- (597.14,275.96) ;
\draw [fill={rgb, 255:red, 80; green, 227; blue, 194 }  ,fill opacity=0.64 ]   (589.53,240.95) -- (589.53,280.74) ;
\draw [fill={rgb, 255:red, 80; green, 227; blue, 194 }  ,fill opacity=0.64 ]   (612.35,226.63) -- (612.35,266.41) ;
\draw [fill={rgb, 255:red, 80; green, 227; blue, 194 }  ,fill opacity=0.64 ]   (604.75,231.4) -- (604.75,271.19) ;
\draw [fill={rgb, 255:red, 80; green, 227; blue, 194 }  ,fill opacity=0.64 ]   (581.93,252.89) -- (612.35,234.59) ;
\draw [fill={rgb, 255:red, 80; green, 227; blue, 194 }  ,fill opacity=0.64 ]   (581.93,260.84) -- (612.35,242.54) ;
\draw [fill={rgb, 255:red, 80; green, 227; blue, 194 }  ,fill opacity=0.64 ]   (581.93,268.8) -- (612.35,250.5) ;
\draw [fill={rgb, 255:red, 80; green, 227; blue, 194 }  ,fill opacity=0.64 ]   (581.93,276.76) -- (612.35,258.46) ;
\draw [fill={rgb, 255:red, 80; green, 227; blue, 194 }  ,fill opacity=0.64 ][line width=2.25]  [dash pattern={on 2.53pt off 3.02pt}]  (624.23,239.47) -- (651.61,221.96) ;
\draw [fill={rgb, 255:red, 80; green, 227; blue, 194 }  ,fill opacity=0.64 ]   (628.51,161.8) -- (543.9,212.91) ;
\draw [fill={rgb, 255:red, 80; green, 227; blue, 194 }  ,fill opacity=0.64 ]   (628.51,161.8) -- (666.53,194.02) ;
\draw [fill={rgb, 255:red, 80; green, 227; blue, 194 }  ,fill opacity=0.64 ][line width=2.25]  [dash pattern={on 2.53pt off 3.02pt}]  (606.84,203.06) -- (634.22,185.55) ;
\draw [fill={rgb, 255:red, 80; green, 227; blue, 194 }  ,fill opacity=0.64 ]   (551.51,208.73) -- (589.53,240.95) ;
\draw [fill={rgb, 255:red, 80; green, 227; blue, 194 }  ,fill opacity=0.64 ]   (574.32,194.41) -- (612.35,226.63) ;
\draw [fill={rgb, 255:red, 80; green, 227; blue, 194 }  ,fill opacity=0.64 ]   (566.72,199.18) -- (604.75,231.4) ;
\draw [fill={rgb, 255:red, 80; green, 227; blue, 194 }  ,fill opacity=0.64 ]   (559.11,203.96) -- (597.14,236.18) ;
\draw [fill={rgb, 255:red, 80; green, 227; blue, 194 }  ,fill opacity=0.64 ]   (551.51,219.47) -- (581.93,201.16) ;
\draw [fill={rgb, 255:red, 80; green, 227; blue, 194 }  ,fill opacity=0.64 ]   (559.11,225.83) -- (589.53,207.53) ;
\draw [fill={rgb, 255:red, 80; green, 227; blue, 194 }  ,fill opacity=0.64 ]   (566.72,232.2) -- (597.14,213.9) ;
\draw [fill={rgb, 255:red, 80; green, 227; blue, 194 }  ,fill opacity=0.64 ]   (574.32,238.56) -- (604.75,220.26) ;

\draw  [fill={rgb, 255:red, 245; green, 166; blue, 35 }  ,fill opacity=1 ] (429.93,223.72) -- (429.93,223.72) -- (437.33,228) -- (467.33,210.68) -- (459.93,206.4) -- cycle ;
\draw  [fill={rgb, 255:red, 245; green, 166; blue, 35 }  ,fill opacity=1 ] (467.33,210.68) -- (467.33,220.68) -- (437.33,238) -- (437.33,228) -- cycle ;
\draw  [fill={rgb, 255:red, 245; green, 166; blue, 35 }  ,fill opacity=1 ] (437.33,228) -- (437.33,238.26) -- (429.93,233.98) -- (429.93,223.72) -- cycle ;
\draw [fill={rgb, 255:red, 245; green, 166; blue, 35 }  ,fill opacity=1 ]   (429.93,223.72) -- (459.93,206.4) ;
\draw [fill={rgb, 255:red, 245; green, 166; blue, 35 }  ,fill opacity=1 ]   (457.33,216.9) -- (457.33,227.16) ;
\draw [fill={rgb, 255:red, 245; green, 166; blue, 35 }  ,fill opacity=1 ]   (447.11,222.56) -- (447.11,232.82) ;
\draw [fill={rgb, 255:red, 245; green, 166; blue, 35 }  ,fill opacity=1 ]   (449.93,212.63) -- (457.33,216.9) ;
\draw [fill={rgb, 255:red, 245; green, 166; blue, 35 }  ,fill opacity=1 ]   (439.71,218.29) -- (447.11,222.56) ;

\draw  [fill={rgb, 255:red, 65; green, 117; blue, 5 }  ,fill opacity=1 ] (319.84,172.21) -- (360.4,199.09) -- (352.29,204.5) -- (352.29,204.5) -- (311.72,177.62) -- cycle ;
\draw  [fill={rgb, 255:red, 65; green, 117; blue, 5 }  ,fill opacity=1 ] (311.72,224.43) -- (311.72,177.62) -- (352.29,204.47) -- (352.29,251.28) -- cycle ;
\draw  [fill={rgb, 255:red, 65; green, 117; blue, 5 }  ,fill opacity=1 ] (360.4,245.88) -- (352.29,251.28) -- (352.29,204.5) -- (360.4,199.09) -- cycle ;
\draw [fill={rgb, 255:red, 65; green, 117; blue, 5 }  ,fill opacity=1 ]   (352.29,224.94) -- (360.4,219.54) ;
\draw [fill={rgb, 255:red, 65; green, 117; blue, 5 }  ,fill opacity=1 ]   (352.29,234.31) -- (360.4,228.9) ;
\draw [fill={rgb, 255:red, 65; green, 117; blue, 5 }  ,fill opacity=1 ]   (352.29,215.58) -- (360.4,210.18) ;
\draw [fill={rgb, 255:red, 65; green, 117; blue, 5 }  ,fill opacity=1 ]   (319.84,183.05) -- (327.95,177.64) ;
\draw [fill={rgb, 255:red, 65; green, 117; blue, 5 }  ,fill opacity=1 ]   (327.95,188.55) -- (336.06,183.15) ;
\draw [fill={rgb, 255:red, 65; green, 117; blue, 5 }  ,fill opacity=1 ]   (336.06,193.51) -- (344.17,188.1) ;
\draw [fill={rgb, 255:red, 65; green, 117; blue, 5 }  ,fill opacity=1 ]   (344.17,198.85) -- (352.29,193.45) ;
\draw [fill={rgb, 255:red, 65; green, 117; blue, 5 }  ,fill opacity=1 ]   (344.17,198.85) -- (344.17,246.11) ;
\draw [fill={rgb, 255:red, 65; green, 117; blue, 5 }  ,fill opacity=1 ]   (336.06,193.51) -- (336.06,240.76) ;
\draw [fill={rgb, 255:red, 65; green, 117; blue, 5 }  ,fill opacity=1 ]   (327.95,188.55) -- (327.95,235.81) ;
\draw [fill={rgb, 255:red, 65; green, 117; blue, 5 }  ,fill opacity=1 ]   (319.84,183.05) -- (319.84,230.3) ;
\draw [fill={rgb, 255:red, 65; green, 117; blue, 5 }  ,fill opacity=1 ]   (311.72,188.73) -- (352.29,215.58) ;
\draw [fill={rgb, 255:red, 65; green, 117; blue, 5 }  ,fill opacity=1 ]   (311.72,215.98) -- (352.29,242.84) ;
\draw [fill={rgb, 255:red, 65; green, 117; blue, 5 }  ,fill opacity=1 ]   (311.72,207.45) -- (352.29,234.31) ;
\draw [fill={rgb, 255:red, 65; green, 117; blue, 5 }  ,fill opacity=1 ]   (311.72,198.09) -- (352.29,224.94) ;
\draw [fill={rgb, 255:red, 65; green, 117; blue, 5 }  ,fill opacity=1 ]   (352.29,242.84) -- (360.4,237.43) ;

\draw  [fill={rgb, 255:red, 208; green, 2; blue, 27 }  ,fill opacity=1 ] (311.72,177.83) -- (352.29,204.71) -- (344.17,210.12) -- (344.17,210.12) -- (303.61,183.23) -- cycle ;
\draw  [fill={rgb, 255:red, 208; green, 2; blue, 27 }  ,fill opacity=1 ] (303.61,230.05) -- (303.61,183.24) -- (344.17,210.09) -- (344.17,256.89) -- cycle ;
\draw  [fill={rgb, 255:red, 208; green, 2; blue, 27 }  ,fill opacity=1 ] (352.29,251.49) -- (344.17,256.9) -- (344.17,210.12) -- (352.29,204.71) -- cycle ;
\draw [fill={rgb, 255:red, 208; green, 2; blue, 27 }  ,fill opacity=1 ]   (344.17,230.56) -- (352.29,225.15) ;
\draw [fill={rgb, 255:red, 208; green, 2; blue, 27 }  ,fill opacity=1 ]   (344.17,239.92) -- (352.29,234.52) ;
\draw [fill={rgb, 255:red, 208; green, 2; blue, 27 }  ,fill opacity=1 ]   (344.17,221.2) -- (352.29,215.79) ;
\draw [fill={rgb, 255:red, 208; green, 2; blue, 27 }  ,fill opacity=1 ]   (311.72,188.67) -- (319.84,183.26) ;
\draw [fill={rgb, 255:red, 208; green, 2; blue, 27 }  ,fill opacity=1 ]   (319.84,194.17) -- (327.95,188.77) ;
\draw [fill={rgb, 255:red, 208; green, 2; blue, 27 }  ,fill opacity=1 ]   (327.95,199.12) -- (336.06,193.72) ;
\draw [fill={rgb, 255:red, 208; green, 2; blue, 27 }  ,fill opacity=1 ]   (336.06,204.47) -- (344.17,199.06) ;
\draw    (336.06,204.47) -- (336.06,251.72) ;
\draw    (327.95,199.12) -- (327.95,246.38) ;
\draw    (319.84,194.17) -- (319.84,241.43) ;
\draw    (311.72,188.67) -- (311.72,235.92) ;
\draw    (303.61,194.35) -- (344.17,221.2) ;
\draw    (303.61,221.6) -- (344.17,248.45) ;
\draw    (303.61,213.07) -- (344.17,239.92) ;
\draw    (303.61,203.71) -- (344.17,230.56) ;
\draw [fill={rgb, 255:red, 208; green, 2; blue, 27 }  ,fill opacity=1 ]   (344.17,248.45) -- (352.29,243.05) ;

\draw  [fill={rgb, 255:red, 245; green, 166; blue, 35 }  ,fill opacity=1 ] (303.61,183.45) -- (344.17,210.33) -- (336.06,215.73) -- (336.06,215.73) -- (295.5,188.85) -- cycle ;
\draw  [fill={rgb, 255:red, 245; green, 166; blue, 35 }  ,fill opacity=1 ] (295.5,235.66) -- (295.5,188.86) -- (336.06,215.7) -- (336.06,262.51) -- cycle ;
\draw  [fill={rgb, 255:red, 245; green, 166; blue, 35 }  ,fill opacity=1 ] (344.17,257.11) -- (336.06,262.52) -- (336.06,215.73) -- (344.17,210.33) -- cycle ;
\draw [fill={rgb, 255:red, 245; green, 166; blue, 35 }  ,fill opacity=1 ]   (336.06,236.18) -- (344.17,230.77) ;
\draw [fill={rgb, 255:red, 245; green, 166; blue, 35 }  ,fill opacity=1 ]   (336.06,245.54) -- (344.17,240.13) ;
\draw [fill={rgb, 255:red, 245; green, 166; blue, 35 }  ,fill opacity=1 ]   (336.06,226.82) -- (344.17,221.41) ;
\draw [fill={rgb, 255:red, 245; green, 166; blue, 35 }  ,fill opacity=1 ]   (303.61,194.28) -- (311.72,188.88) ;
\draw [fill={rgb, 255:red, 245; green, 166; blue, 35 }  ,fill opacity=1 ]   (311.72,199.79) -- (319.84,194.38) ;
\draw [fill={rgb, 255:red, 245; green, 166; blue, 35 }  ,fill opacity=1 ]   (319.84,204.74) -- (327.95,199.34) ;
\draw [fill={rgb, 255:red, 245; green, 166; blue, 35 }  ,fill opacity=1 ]   (327.95,210.09) -- (336.06,204.68) ;
\draw [fill={rgb, 255:red, 245; green, 166; blue, 35 }  ,fill opacity=1 ]   (327.95,210.09) -- (327.95,257.34) ;
\draw [fill={rgb, 255:red, 245; green, 166; blue, 35 }  ,fill opacity=1 ]   (319.84,204.74) -- (319.84,252) ;
\draw [fill={rgb, 255:red, 245; green, 166; blue, 35 }  ,fill opacity=1 ]   (311.72,199.79) -- (311.72,247.04) ;
\draw [fill={rgb, 255:red, 245; green, 166; blue, 35 }  ,fill opacity=1 ]   (303.61,194.28) -- (303.61,241.54) ;
\draw [fill={rgb, 255:red, 245; green, 166; blue, 35 }  ,fill opacity=1 ]   (295.5,199.96) -- (336.06,226.82) ;
\draw [fill={rgb, 255:red, 245; green, 166; blue, 35 }  ,fill opacity=1 ]   (295.5,227.22) -- (336.06,254.07) ;
\draw [fill={rgb, 255:red, 245; green, 166; blue, 35 }  ,fill opacity=1 ]   (295.5,218.69) -- (336.06,245.54) ;
\draw [fill={rgb, 255:red, 245; green, 166; blue, 35 }  ,fill opacity=1 ]   (295.5,209.32) -- (336.06,236.18) ;
\draw [fill={rgb, 255:red, 245; green, 166; blue, 35 }  ,fill opacity=1 ]   (336.06,254.07) -- (344.17,248.66) ;

\draw  [color={rgb, 255:red, 245; green, 166; blue, 35 }  ,draw opacity=1 ][dash pattern={on 4.5pt off 4.5pt}] (151.81,218.81) .. controls (151.81,193.39) and (172.57,172.79) .. (198.17,172.79) .. controls (223.78,172.79) and (244.53,193.39) .. (244.53,218.81) .. controls (244.53,244.22) and (223.78,264.83) .. (198.17,264.83) .. controls (172.57,264.83) and (151.81,244.22) .. (151.81,218.81) -- cycle ;
\draw  [fill={rgb, 255:red, 245; green, 166; blue, 35 }  ,fill opacity=1 ] (176.96,198.27) -- (203.1,214.22) -- (194.49,219.57) -- (168.34,203.62) -- (168.34,203.62) -- cycle ;
\draw  [fill={rgb, 255:red, 245; green, 166; blue, 35 }  ,fill opacity=1 ] (168.34,229.39) -- (168.34,203.63) -- (194.49,219.57) -- (194.49,245.33) -- cycle ;
\draw [fill={rgb, 255:red, 245; green, 166; blue, 35 }  ,fill opacity=1 ]   (185.9,213.93) -- (185.98,240.13) ;
\draw [fill={rgb, 255:red, 245; green, 166; blue, 35 }  ,fill opacity=1 ]   (177.65,209.18) -- (177.47,234.74) ;
\draw  [fill={rgb, 255:red, 245; green, 166; blue, 35 }  ,fill opacity=1 ] (203.1,239.8) -- (194.49,244.98) -- (194.49,219.4) -- (203.1,214.22) -- cycle ;
\draw    (194.49,228.53) -- (203.1,223.35) ;
\draw    (194.49,237.52) -- (203.1,232.33) ;
\draw    (185.9,213.93) -- (194.52,208.74) ;
\draw    (177,209.69) -- (185.62,204.5) ;
\draw    (168.09,204.12) -- (176.71,198.77) ;
\draw    (168.34,221.57) -- (194.49,237.52) ;
\draw    (168.34,212.59) -- (194.49,228.53) ;
\draw    (176.71,198.77) -- (202.85,214.71) ;

\draw  [color={rgb, 255:red, 245; green, 166; blue, 35 }  ,draw opacity=1 ][dash pattern={on 4.5pt off 4.5pt}] (405.15,221.75) .. controls (405.15,197.47) and (424.83,177.79) .. (449.11,177.79) .. controls (473.39,177.79) and (493.07,197.47) .. (493.07,221.75) .. controls (493.07,246.03) and (473.39,265.71) .. (449.11,265.71) .. controls (424.83,265.71) and (405.15,246.03) .. (405.15,221.75) -- cycle ;
\draw  [fill={rgb, 255:red, 74; green, 144; blue, 226 }  ,fill opacity=1 ] (70.34,163.57) -- (101.6,181.46) -- (80.31,193.5) -- (80.31,193.5) -- (49.05,175.69) -- cycle ;
\draw  [fill={rgb, 255:red, 74; green, 144; blue, 226 }  ,fill opacity=1 ] (24.65,228.09) -- (24.65,161.8) -- (80.31,193.5) -- (80.31,259.78) -- cycle ;
\draw [fill={rgb, 255:red, 74; green, 144; blue, 226 }  ,fill opacity=1 ]   (80.31,252.57) -- (24.65,220.87) ;
\draw [fill={rgb, 255:red, 74; green, 144; blue, 226 }  ,fill opacity=1 ]   (80.31,242.72) -- (24.65,211.02) ;
\draw [fill={rgb, 255:red, 74; green, 144; blue, 226 }  ,fill opacity=1 ]   (80.31,203.34) -- (24.65,171.65) ;
\draw [fill={rgb, 255:red, 74; green, 144; blue, 226 }  ,fill opacity=1 ]   (80.31,213.19) -- (24.65,181.49) ;
\draw [fill={rgb, 255:red, 74; green, 144; blue, 226 }  ,fill opacity=1 ]   (80.31,223.03) -- (24.65,191.33) ;
\draw [fill={rgb, 255:red, 74; green, 144; blue, 226 }  ,fill opacity=1 ]   (80.31,232.88) -- (24.65,201.18) ;

\draw [fill={rgb, 255:red, 74; green, 144; blue, 226 }  ,fill opacity=1 ]   (73.45,189.59) -- (73.45,255.87) ;
\draw [fill={rgb, 255:red, 74; green, 144; blue, 226 }  ,fill opacity=1 ]   (32.79,166.43) -- (32.79,232.72) ;
\draw [fill={rgb, 255:red, 74; green, 144; blue, 226 }  ,fill opacity=1 ]   (41.54,171.42) -- (41.54,237.7) ;
\draw [fill={rgb, 255:red, 74; green, 144; blue, 226 }  ,fill opacity=1 ]   (49.05,175.69) -- (49.05,241.98) ;
\draw [fill={rgb, 255:red, 74; green, 144; blue, 226 }  ,fill opacity=1 ]   (57.18,180.33) -- (57.18,246.61) ;
\draw [fill={rgb, 255:red, 74; green, 144; blue, 226 }  ,fill opacity=1 ]   (65.31,184.96) -- (65.31,251.24) ;

\draw  [fill={rgb, 255:red, 74; green, 144; blue, 226 }  ,fill opacity=1 ] (101.6,247.73) -- (80.31,259.85) -- (80.31,193.59) -- (101.6,181.46) -- cycle ;
\draw [fill={rgb, 255:red, 74; green, 144; blue, 226 }  ,fill opacity=1 ]   (87.41,189.55) -- (87.41,255.81) ;
\draw [fill={rgb, 255:red, 74; green, 144; blue, 226 }  ,fill opacity=1 ]   (94.5,185.5) -- (94.5,251.77) ;
\draw [fill={rgb, 255:red, 74; green, 144; blue, 226 }  ,fill opacity=1 ]   (80.31,232.88) -- (101.6,220.76) ;
\draw [fill={rgb, 255:red, 74; green, 144; blue, 226 }  ,fill opacity=1 ]   (80.31,242.72) -- (101.6,230.6) ;
\draw [fill={rgb, 255:red, 74; green, 144; blue, 226 }  ,fill opacity=1 ]   (80.31,223.03) -- (101.6,210.91) ;
\draw [fill={rgb, 255:red, 74; green, 144; blue, 226 }  ,fill opacity=1 ]   (80.31,252.57) -- (101.6,240.44) ;
\draw [fill={rgb, 255:red, 74; green, 144; blue, 226 }  ,fill opacity=1 ]   (80.31,213.19) -- (101.6,201.07) ;
\draw [fill={rgb, 255:red, 74; green, 144; blue, 226 }  ,fill opacity=1 ]   (80.31,203.34) -- (101.6,191.22) ;

\draw    (24.65,161.8) -- (80.31,193.5) ;
\draw [fill={rgb, 255:red, 74; green, 144; blue, 226 }  ,fill opacity=1 ]   (45.94,149.77) -- (101.6,181.46) ;
\draw [fill={rgb, 255:red, 74; green, 144; blue, 226 }  ,fill opacity=1 ]   (24.65,161.8) -- (45.94,149.68) ;
\draw [fill={rgb, 255:red, 74; green, 144; blue, 226 }  ,fill opacity=1 ]   (31.75,157.85) -- (87.41,189.55) ;
\draw [fill={rgb, 255:red, 74; green, 144; blue, 226 }  ,fill opacity=1 ]   (38.84,153.81) -- (94.5,185.5) ;
\draw [fill={rgb, 255:red, 74; green, 144; blue, 226 }  ,fill opacity=1 ]   (41.54,171.42) -- (62.83,159.3) ;
\draw [fill={rgb, 255:red, 74; green, 144; blue, 226 }  ,fill opacity=1 ]   (49.05,175.69) -- (70.34,163.57) ;
\draw [fill={rgb, 255:red, 74; green, 144; blue, 226 }  ,fill opacity=1 ]   (32.79,166.43) -- (54.07,154.31) ;

\draw [fill={rgb, 255:red, 74; green, 144; blue, 226 }  ,fill opacity=1 ]   (57.18,180.33) -- (78.47,168.2) ;
\draw [fill={rgb, 255:red, 74; green, 144; blue, 226 }  ,fill opacity=1 ]   (65.31,184.96) -- (86.6,172.83) ;

\draw    (80.31,193.59) -- (101.6,181.46) ;
\draw    (73.45,189.59) -- (94.73,177.47) ;
\draw  [fill={rgb, 255:red, 65; green, 117; blue, 5 }  ,fill opacity=0.89 ] (45.94,149.77) -- (70.34,163.57) -- (63.34,167.4) -- (63.34,167.4) -- (38.84,153.81) -- cycle ;
\draw  [fill={rgb, 255:red, 208; green, 2; blue, 27 }  ,fill opacity=0.85 ] (38.84,153.81) -- (63.34,167.4) -- (56.15,171.36) -- (31.75,157.85) -- (31.75,157.85) -- cycle ;
\draw  [fill={rgb, 255:red, 245; green, 166; blue, 35 }  ,fill opacity=0.8 ] (31.75,157.85) -- (56.15,171.36) -- (49.05,175.69) -- (24.65,161.8) -- (24.65,161.8) -- cycle ;
\draw  [fill={rgb, 255:red, 245; green, 166; blue, 35 }  ,fill opacity=0.81 ] (24.65,161.8) -- (49.05,175.69) -- (48.97,205.01) -- (24.65,191.33) -- (24.65,191.33) -- cycle ;

\draw [color={rgb, 255:red, 245; green, 166; blue, 35 }  ,draw opacity=1 ][line width=2.25]    (118.48,218.14) -- (151.81,218.81) ;
\draw [color={rgb, 255:red, 245; green, 166; blue, 35 }  ,draw opacity=1 ][line width=2.25]    (371.81,221.08) -- (405.15,221.75) ;
\draw [color={rgb, 255:red, 245; green, 166; blue, 35 }  ,draw opacity=1 ][line width=2.25]    (493.07,221.75) -- (521.4,222.31) ;
\draw [shift={(526.4,222.41)}, rotate = 181.15] [fill={rgb, 255:red, 245; green, 166; blue, 35 }  ,fill opacity=1 ][line width=0.08]  [draw opacity=0] (10,-4.8) -- (0,0) -- (10,4.8) -- cycle    ;
\draw [color={rgb, 255:red, 245; green, 166; blue, 35 }  ,draw opacity=1 ][line width=2.25]    (244.53,218.81) -- (272.87,219.37) ;
\draw [shift={(277.87,219.47)}, rotate = 181.15] [fill={rgb, 255:red, 245; green, 166; blue, 35 }  ,fill opacity=1 ][line width=0.08]  [draw opacity=0] (10,-4.8) -- (0,0) -- (10,4.8) -- cycle    ;

\draw (197,244) node [anchor=north west][inner sep=0.75pt]   [align=left] {{\scriptsize $1$}};
\draw (222,228) node [anchor=north west][inner sep=0.75pt]   [align=left] {{\scriptsize $1$}};
\draw (209,235) node [anchor=north west][inner sep=0.75pt]   [align=left] {{\scriptsize $1$}};
\draw (90,253) node [anchor=north west][inner sep=0.75pt]   [align=left] {{\scriptsize 3}};
\draw (171.72,237) node [anchor=north west][inner sep=0.75pt]   [align=left] {{\scriptsize 3}};
\draw (154,213) node [anchor=north west][inner sep=0.75pt]   [align=left] {{\scriptsize 3}};
\draw (10,187) node [anchor=north west][inner sep=0.75pt]   [align=left] {{\scriptsize 7}};
\draw (43.54,244) node [anchor=north west][inner sep=0.75pt]   [align=left] {{\scriptsize 7}};
\draw (552,266) node [anchor=north west][inner sep=0.75pt]   [align=left] {{\scriptsize 5}};
\draw (530,226.5) node [anchor=north west][inner sep=0.75pt]   [align=left] {{\scriptsize 5}};
\draw (620.57,260) node [anchor=north west][inner sep=0.75pt]   [align=left] {{\scriptsize 128}};
\draw (346.17,253) node [anchor=north west][inner sep=0.75pt]   [align=left] {{\scriptsize 3}};
\draw (280,207) node [anchor=north west][inner sep=0.75pt]   [align=left] {{\scriptsize 5}};
\draw (305.61,247) node [anchor=north west][inner sep=0.75pt]   [align=left] {{\scriptsize 5}};
\draw (451.11,229) node [anchor=north west][inner sep=0.75pt]   [align=left] {{\scriptsize 3}};
\draw (415,220) node [anchor=north west][inner sep=0.75pt]   [align=left] {{\scriptsize $1$}};
\draw (424,240) node [anchor=north west][inner sep=0.75pt]   [align=left] {{\scriptsize $1$}};
\draw (85.33,124) node [anchor=north west][inner sep=0.75pt]   [align=left] {\scriptsize Depthwise Convolution Operation};
\draw (380,124) node [anchor=north west][inner sep=0.75pt]   [align=left] {\scriptsize Pointwise Convolution Operation};
\draw (429,156) node [anchor=north west][inner sep=0.75pt]   [align=left] {{\scriptsize x 128}};
\end{tikzpicture}
    \caption{A 3D illustration of the overall process of depthwise separable convolution. A standard convolution will transform the image 128 times whereas a depthwise separable convolution will transform the image only once and then the transformed image is elongated to 128 channels which allows the neural network to process more data whilst utilizing fewer FLOPs resulting  in  an unprecedented  computational  efficiency.}
    \label{fig}
\end{figure}

This approach results in to more efficient computation requiring $7.3\%$ fewer FLOPs and $11.5\%$ fewer trainable parameters than the original grouped convolutions at the cost of $1.17\%$ loss in Top-1 accuracy. Depthwise separable convolutions have been proven in reducing the trained model size and computational resources, both in theory and in practice \cite{b27,b34}.

The second half of the training utilizes a widely used model compression technique which has further significant impact, both in computational efficiency at training time and on the final trained model size. We present details on our approach below.
\newline

\subsection{Model Compression}

Model compression is one the many popular methods of reducing complexity of a DNN to make it more computationally efficient by discarding redundant elements that do not influence performance of the model.
\newline
\textbf{Group-wise Pruning:}
The goal of group-wise pruning is to discard inconsequential filters for every group $g$ during the training stage which is determined by the $L_{1}$-Normalization of $X^{g_{ij}}$ where for every group $g$, $i$ is the input and $j$ is the output. We introduce pruning hyper-parameter $p$ and set it to $4$ which allows the network to determine the number of filters to be removed before the first stage of depthwise separable convolution during the training process. We also add a class balanced focal loss function \cite{b8} to smoothen the pruning process.
\medskip
\newline
\textbf{Cardinality:}
We introduce a new dimension to the network called \textit{Cardinality} denoted by ${C}$ in addition to the existing depth and width of the network in order to minimize the loss in accuracy during the pruning process. We set the the cardinality $C$ equal to the number of output channels because the number was found to be very small for division. Experiments demonstrate that increasing cardinality is a more effective way of gaining accuracy than going deeper or wider, especially when depth and width starts to give diminishing returns \cite{b5}.
\vskip 0.06in
Consider a group convolution comprised of $G$ groups of size $H \times H \times C_{I} \times C_{O}$ where $C_{I}=$\( \frac{I}{G} \) and $C_{O}=$\( \frac{O}{G} \), the total number of inconsequential filters that will be pruned before the first stage of depthwise separable convolution during the training process can be determined as follows:
\vskip 0.05in
\begin{equation}
\label{eqn_4}
G \cdot C_{x}=I \cdot C-p \cdot I
\end{equation}
\vskip 0.05in

This approach further results in a significant decrease in computational resources requiring $60.99\%$ fewer FLOPs and $65.21\%$ fewer trainable parameters at the cost of $0.88\%$ loss in Top-1 accuracy for CIFAR-10 dataset.
\vskip 0.1in

\begin{table*}[t]
\centering
\caption{Comparison of Performance}
\begin{tabular}{|l|c|c|c|c|} 
\hline
\rowcolor[rgb]{0.929,0.929,0.929} \multicolumn{1}{|c|}{Architecture} & FLOPs (in millions) & Parameters (in millions) & CIFAR-10 Top-1 Accuracy & \multicolumn{1}{l|}{Trained Model Size} \\ 
\hline
CondenseNet (Baseline Architecture) & 65.81 & 0.52 & 94.24\% & 16.7 MB \\ 
\hline
CondenseNeXt (Our Proposed~Architecture) & 23.80 & 0.16 & 92.28\% & 2.9 MB \\
\hline
\end{tabular}
\vskip 0.06in
Table I provides a comparison between CondenseNet (the baseline architecture) vs. CondenseNeXt (the proposed architecture in this paper) in terms of performance for 200 epochs each utilizing the training infrastructure as outlined in section 5 in this paper.
\end{table*}

\subsection{Activation Function}
\vskip 0.03in
Activation functions in deep neural networks determine the output of a neuron at a given input or a set of inputs by limiting the amplitude of the output. It helps the network understand and learn complex patterns of the input data. Furthermore, non-linear activation functions such as ReLU (Rectified Linear Units) and ReLU6 \cite{b11} enable such networks to perform non-trivial computations with the help of fewer neurons.

Our approach utilizes ReLU6 activation function along with Batch Normalization before each layer. In ReLU6, units are capped at 6 which promotes an earlier learning of sparse features. Furthermore, it prevents the explosion of positive gradients to infinity. ReLU6 is defined as follows:
\vskip 0.05in
\begin{equation}
\label{eqn_5}
f(x) = min(max(0,x),6)
\end{equation}
\vskip 0.05in

Our experiments show that this approach was able to achieve a small gain of $0.577\%$ in Top-1 accuracy for CIFAR-10 dataset after implementing the model compression technique as described earlier in this section.

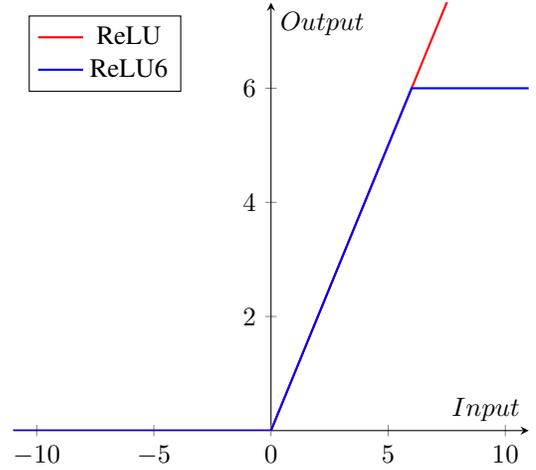
\begin{figure}
\centering
\begin{tikzpicture}
\begin{axis}[
    legend pos=north west,
    axis lines=middle,
    extra x ticks=0,
    xmax=11,
    xmin=-11,
    ymin=0,
    ymax=7.5,
    xlabel={$Input$},
    ylabel={$Output$}]
\addplot [domain=-11:11, samples=500, thick, red] {max(0,x)};
\addlegendentry{ReLU}
\addplot [domain=-11:11, samples=1000, thick, blue] {min(max(0,x),6)};
\addlegendentry{ReLU6}
\end{axis}
\end{tikzpicture}
\caption{Graph demonstrating the difference between ReLU and ReLU6 activation functions.}
\end{figure}

\section{Training Infrastructure}
\medskip
\begin{itemize}
\setlength\itemsep{0.35em}
\item Intel Xeon Gold 6126 12-core CPU with 32 GB RAM.
\item NVIDIA Tesla V100 GPU.
\item CUDA Toolkit 10.1.243.
\item Python version 3.7.9.
\item PyTorch version 1.1.0.
\item PyCharm Professional IDE version 2019.3.5.
\end{itemize}

\medskip

This cyberinfrastructure is provided and managed by the Research Technologies division at the Indiana University which supported our work in part by Shared University Research grants from IBM Inc. to Indiana University and Lilly Endowment Inc. through its support for the Indiana University Pervasive Technology Institute \cite{b22}.

\medskip

\section{Experiment and Results}

Training results presented in this report are based on the evaluation of our proposed CondenseNeXt architecture on CIFAR-10 image classification dataset.\\
\newline
\textbf{Training Details:} Our network was implemented using the PyTorch framework and trained on NVIDIA's Tesla V100 GPU  with Nesterov Momentum Weight of 0.9 and Stochastic Gradient Descent (SGD) for 200 epochs. Furthermore, the training utilizes cosine shape learning rate and dropout rate of 0.1.
\bigskip
\newline
\textbf{Dataset:} CIFAR-10 dataset \cite{b11,b7}, first introduced by Alex Krizhevsky in \cite{b1}, is one of the most popular datasets used in the field of machine learning research. It comprises of 60,000 RGB images of 10 different classes of size 32$ \times $32 pixels split into two sets of 50,000 for training and 10,000 for testing. We choose to use standard data augmentation scheme \cite{b3} successfully implemented in the baseline architecture.
\smallskip
\newline
\textbf{Training Results:} Our model was evaluated with a single crop of inputs on CIFAR-10 dataset for 200 epochs. Table I provides a comparison between the baseline architecture and our proposed architecture in terms of FLOPs, parameters, Top-1 accuracy and trained model size. Table II provides a comparison of training time between the two architectures and figure 3 plots the accuracy of our proposed model. The Top-1 Accuracy of our proposed architecture is $92.28\%$ with a trained model size of 2.9 MB.

\begin{table}
\centering
\caption{Comparison of Training Time}
\resizebox{\linewidth}{!}{%
\begin{tabular}{|>{\hspace{0pt}}m{0.24\linewidth}|>{\centering\hspace{0pt}}m{0.296\linewidth}|>{\centering\arraybackslash\hspace{0pt}}m{0.392\linewidth}|} 
\hline
\rowcolor[rgb]{0.929,0.929,0.929} \multicolumn{1}{|>{\centering\hspace{0pt}}m{0.24\linewidth}|}{Architecture} & Total Training Time\par{}(HH:MM:SS) & Training Time Per Epoch \par{}(in seconds) \\ 
\hline
CondenseNet & 02:59:19 & 53.79 \\ 
\hline
CondenseNeXt & 01:04:48 & 19.42 \\
\hline
\end{tabular}
}
\vskip 0.06in
Table II provides a comparison of training time between CondenseNet (the baseline architecture) vs. CondenseNeXt (the proposed architecture in this paper) for 200 epochs each, utilizing the training infrastructure and setup as outlined in sections 5 and 6 in this paper.
\end{table}

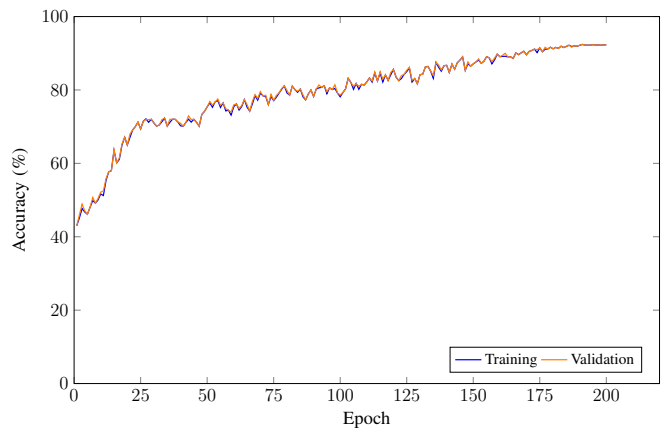
\begin{figure}
 \begin{tikzpicture}[scale=0.58]

\begin{axis}
[
legend pos=south east,
xlabel={Epoch}, ylabel={Accuracy (\%)}, xmin=0, xmax=220,
ymin=0.0, ymax=100,
height=10cm,
width=15cm,
legend columns=5,
ticklabel style = {font=\large, thick},
label style = {font=\large},
xtick={0,25,50,75,100,125,150,175, 200},
ytick={0,20,40,60,80, 100}
]
\addplot
[
line width=.8pt,
color=blue,
]
coordinates{
(1,     43.01)
(2,     45.17)
(3,     47.68)
(4,     46.74) 
(5,     46.14)
(6,     48.03) 
(7,     49.87)
(8,     49.16) 
(9,     50.03) 
(10,    51.65) 
(11,    51.23)
(12,    55.35) 
(13,    57.7)
(14,    58.00) 
(15,    63.56)
(16,    60.12) 
(17,    61.08) 
(18,    65.01)
(19,    67.11)
(20,    65.00)
(21,	67.09)
(22,	69.11)
(23,	70.0)
(24,	71.23)
(25,	69.39)
(26,	71.47)
(27,	72.13)
(28,	71.2)
(29	,   72.01)
(30	,   71.00)
(31	,   70.14)
(32	,   70.45)
(33	,   71.34)
(34	,   72.32)
(35	,   70.16)
(36	,   71.19)
(37	,   72.03)
(38	,   72)
(39	,   71.27)
(40	,   70.24)
(41,	70.1)
(42,	71.14)
(43,	72.04)
(44,	71.23)
(45,	72)
(46,	71.19)
(47,	70.12)
(48,	73.34)
(49,	74.16)
(50,	75.38)
(51,	76.47)
(52,	75.27)
(53,	76.77)
(54,	77.14)
(55,	75.24)
(56,	76.45)
(57,	74.29)
(58,	74.5)
(59,	73.15)
(60,	75.43)
(61,	76.2)
(62,	74.53)
(63,	75.49)
(64,	77.38)
(65,	75.29)
(66,	74.24)
(67,	76.2)
(68,	78.43)
(69,	77.22)
(70,	79.14)
(71,	78.38)
(72,	78.25)
(73,	76.03)
(74,	78.19)
(75,	77.1)
(76,	78.0)
(77,	79.12)
(78,	80.19)
(79,	81.12)
(80,	79.18)
(81,	78.58)
(82,	81.03)
(83,	80.07)
(84,	79.35)
(85,	80.25)
(86,	78.16)
(87,	77.28)
(88,	78.9)
(89,	79.99)
(90,	78.22)
(91,	80.24)
(92,	80.57)
(93,	80.7)
(94,	81.13)
(95,	78.99)
(96,	80.57)
(97,	80.12)
(98,	80.41)
(99,	79.13)
(100,	78.14)
(101,	79.27)
(102,	80.28)
(103,	83.17)
(104,	82.07)
(105,	80.17)
(106,	81.76)
(107,	80.18)
(108,	81.54)
(109,	81.34)
(110,	82.19)
(111,	83.28)
(112,	82.14)
(113,	84.62)
(114,	82.43)
(115,	84.61)
(116,	82.17)
(117,	84.09)
(118,	82.6)
(119,	84.11)
(120,	85.6)
(121,	83.5)
(122,	82.53)
(123,	83.11)
(124,	84.23)
(125,	85.0)
(126,	86.0)
(127,	82.16)
(128,	83.11)
(129,	81.69)
(130,	84.17)
(131,	84.29)
(132,	86.26)
(133,	86.41)
(134,	85.09)
(135,	83.18)
(136,	87.52)
(137,	86.16)
(138,	85.11)
(139,	86.62)
(140,	86.71)
(141,	84.76)
(142,	87.06)
(143,	85.65)
(144,	87.41)
(145,	88.13)
(146,	89.01)
(147,	85.35)
(148,	87.14)
(149,	86.47)
(150,	87.1)
(151,	87.62)
(152,	88.16)
(153,	87.27)
(154,	87.81)
(155,	89.05)
(156,	88.76)
(157,	87.12)
(158,	88.19)
(159,	89.75)
(160,	89.06)
(161,	89.14)
(162,	89.18)
(163,	89.02)
(164,	89.05)
(165,	88.6)
(166,	90.11)
(167,	89.61)
(168,	90.11)
(169,	90.55)
(170,	89.58)
(171,	90.5)
(172,	90.76)
(173,	91.13)
(174,	90.18)
(175,	91.45)
(176,	90.41)
(177,	91.11)
(178,	91.16)
(179,	91.65)
(180,	91.21)
(181,	91.6)
(182,	91.34)
(183,	91.92)
(184,	91.56)
(185,	91.87)
(186,	92.19)
(187,	91.84)
(188,	92)
(189,	91.91)
(190,	92.17)
(191,	92.41)
(192,	92.22)
(193,	92.21)
(194,	92.21)
(195,	92.34)
(196,	92.23)
(197,	92.26)
(198,	92.21)
(199,	92.29)
(200,	92.28)
    };
\addlegendentry{Training}
\addplot
[
line width=.8pt,
color=orange,
]
coordinates{
(1,     43.01)
(2,     45.97)
(3,     48.87)
(4,     47.04) 
(5,     46.14)
(6,     48.03) 
(7,     50.7)
(8,     49.16) 
(9,     50.43) 
(10,    52.16) 
(11,    52.48)
(12,    55.85) 
(13,    57.7)
(14,    58.00) 
(15,    63.56)
(16,    60.12) 
(17,    61.48) 
(18,    65.31)
(19,    67.11)
(20,    65.00)
(21,	67.99)
(22,	69.11)
(23,	70.0)
(24,	71.23)
(25,	69.39)
(26,	71.47)
(27,	71.93)
(28,	71.92)
(29	,   72.01)
(30	,   71.00)
(31	,   70.14)
(32	,   70.45)
(33	,   71.94)
(34	,   72.32)
(35	,   70.16)
(36	,   71.99)
(37	,   72.03)
(38	,   72)
(39	,   71.27)
(40	,   70.94)
(41,	70.1)
(42,	71.14)
(43,	72.94)
(44,	71.93)
(45,	72)
(46,	71.19)
(47,	70.12)
(48,	73.34)
(49,	74.16)
(50,	75.38)
(51,	76.87)
(52,	75.87)
(53,	76.77)
(54,	77.54)
(55,	75.84)
(56,	76.45)
(57,	74.89)
(58,	74.5)
(59,	73.95)
(60,	75.93)
(61,	76.2)
(62,	74.93)
(63,	75.79)
(64,	77.38)
(65,	75.99)
(66,	74.24)
(67,	76.9)
(68,	78.73)
(69,	77.72)
(70,	79.54)
(71,	78.38)
(72,	78.55)
(73,	76.03)
(74,	78.79)
(75,	77.2)
(76,	78.5)
(77,	79.22)
(78,	80.39)
(79,	81.12)
(80,	79.88)
(81,	78.58)
(82,	81.03)
(83,	80.07)
(84,	79.65)
(85,	80.25)
(86,	78.66)
(87,	77.28)
(88,	78.9)
(89,	79.99)
(90,	78.22)
(91,	80.24)
(92,	81.37)
(93,	80.7)
(94,	81.13)
(95,	79.56)
(96,	80.57)
(97,	80.12)
(98,	81.41)
(99,	79.13)
(100,	78.64)
(101,	79.27)
(102,	80.28)
(103,	83.17)
(104,	82.07)
(105,	80.97)
(106,	81.76)
(107,	80.98)
(108,	81.54)
(109,	81.34)
(110,	82.19)
(111,	83.28)
(112,	82.44)
(113,	84.92)
(114,	82.43)
(115,	84.99)
(116,	82.97)
(117,	84.09)
(118,	82.6)
(119,	84.81)
(120,	85.56)
(121,	83.5)
(122,	82.53)
(123,	83.91)
(124,	84.23)
(125,	85.38)
(126,	86.18)
(127,	82.66)
(128,	83.11)
(129,	81.66)
(130,	84.17)
(131,	84.29)
(132,	86.26)
(133,	86.41)
(134,	85.09)
(135,	83.98)
(136,	87.52)
(137,	86.76)
(138,	85.51)
(139,	86.62)
(140,	86.71)
(141,	84.76)
(142,	87.06)
(143,	85.65)
(144,	87.41)
(145,	88.13)
(146,	89.01)
(147,	85.35)
(148,	87.64)
(149,	86.47)
(150,	87.1)
(151,	87.62)
(152,	88.46)
(153,	87.27)
(154,	87.81)
(155,	89.05)
(156,	88.76)
(157,	87.72)
(158,	88.69)
(159,	89.75)
(160,	89.06)
(161,	89.44)
(162,	89.98)
(163,	89.02)
(164,	89.05)
(165,	88.6)
(166,	90.11)
(167,	89.61)
(168,	90.11)
(169,	90.55)
(170,	89.58)
(171,	90.5)
(172,	90.76)
(173,	91.13)
(174,	90.98)
(175,	91.45)
(176,	90.41)
(177,	91.61)
(178,	91.16)
(179,	91.65)
(180,	91.21)
(181,	91.6)
(182,	91.34)
(183,	91.92)
(184,	91.56)
(185,	91.87)
(186,	92.19)
(187,	91.64)
(188,	92)
(189,	91.91)
(190,	92.17)
(191,	92.41)
(192,	92.22)
(193,	92.21)
(194,	92.21)
(195,	92.34)
(196,	92.23)
(197,	92.26)
(198,	92.21)
(199,	92.29)
(200,	92.28)
    };
\addlegendentry{Validation}
\end{axis}
\end{tikzpicture}
    \caption{Training Overview of CondenseNeXt on CIFAR-10 dataset based on the training infrastructure and setup outlined in sections 5 and 6 of this paper.}
    \label{fig:my_label}
\end{figure}
\section{Conclusion}
\vskip 0.03in
In this paper, we introduce CondenseNeXt: an ultra-efficient convolutional neural network architecture for embedded systems that ameliorates the performance of baseline convolutional neural network architecture, CondenseNet, by utilizing depthwise separable convolution and model compression techniques to significantly reduce computational resources required for training from scratch without a pre-trained model. The primary goal of our work presented in this paper is to reduce the trained model size to less than 3.0 MB so that it can be deployed for real-time inference on embedded systems such as \cite{b81,b82} with constrained computational resources.
\vskip 0.06in
Our work presented in this paper also coincides with the trend that started with Xception and MobileNets that bespeak that any traditional convolution can be replaced by depthwise separable convolution to obtain a trained model that is relatively small in size and efficient in terms of FLOPs which is corroborated by strong theoretical foundations as well as experimental results. We hope that our work further bolsters this trend. Furthermore, we also anticipate the use of depthwise separable convolutions into the design of new deep neural network architectures in the future. As desire and demand for smaller yet efficient deep neural network architectures increase with innovations in, including but not limited to, embedded systems with limited computational resources, the scope for research and innovation in this field is immensurable.
\vskip 0.05in

\section*{Acknowledgment}
The authors acknowledge the Indiana University Pervasive Technology Institute for providing supercomputing and storage resources that have contributed to the research results reported within this paper.

\end{document}